\documentclass[conference]{IEEEtran}
\IEEEoverridecommandlockouts
\usepackage{cite}
\usepackage{amsmath,amssymb,amsfonts}
\usepackage{graphicx}
\usepackage{textcomp}
\usepackage{xcolor}

\usepackage{times}
\usepackage{soul}
\usepackage{url}
\usepackage[hidelinks]{hyperref}
\usepackage[utf8]{inputenc}
\usepackage[small]{caption}
\usepackage{algorithm}
\usepackage[switch]{lineno}

\usepackage[noend]{algpseudocode}
\usepackage{booktabs}
\usepackage{multirow}

\newtheorem{theorem}{Theorem}
\newtheorem{lemma}{Lemma}

\newcommand{\tb}[1]{\textbf{#1}}
\def \D {\mathcal{D}}
\def \N {\mathcal{N}}

\def \R {\mathbb{R}}

\def \tp {\tilde{p}}

\def\BibTeX{{\rm B\kern-.05em{\sc i\kern-.025em b}\kern-.08em
    T\kern-.1667em\lower.7ex\hbox{E}\kern-.125emX}}

\begin{document}

\title{Trustworthy Personalized Bayesian Federated Learning via Posterior Fine-Tune
}

% \author{\IEEEauthorblockN{Anonymous Authors}
% }

\author{
\IEEEauthorblockN{Mengen Luo}
\IEEEauthorblockA{\textit{Tsinghua-Berkeley Shenzhen Institute} \\
\textit{Tsinghua University}\\
ShenZhen, China \\
lme21@mails.tsinghua.edu.cn}
\and
\IEEEauthorblockN{Chi Xu}
\IEEEauthorblockA{\textit{Tsinghua-Berkeley Shenzhen Institute} \\
\textit{Tsinghua University}\\
ShenZhen, China \\
xu-c23@mails.tsinghua.edu.cn}
\and
\IEEEauthorblockN{Ercan Engin Kuruoglu}
\IEEEauthorblockA{\textit{Tsinghua-Berkeley Shenzhen Institute} \\
\textit{Tsinghua University}\\
ShenZhen, China \\
kuruoglu@sz.tsinghua.edu.cn}
}

\maketitle

\begin{abstract}
% Performance degradation due to data heterogeneity and low output interpretability are the most significant challenges faced by federated learning in practical applications. Personalized federated learning diverges from traditional approaches, as it no longer seeks to train a single model but instead tailors a unique personalized model for each client. However, previous works have primarily focused on personalization over deterministic  neural networks, leading to a lack of robustness and interpretability. In this work, we establish a novel framework for personalized Bayesian federated learning, incorporating Bayesian methodology which enhances the algorithm’s ability to quantify uncertainty. Our algorithm exhibits excellent compatibility, rendering it capable of providing uncertainty quantification abilities to numerous ongoing research endeavors. Moreover, we propose a viable method with acceptable computational and storage complexity which achieves personalization by fine-tuning the global posterior using normalizing flows. Finally, we  evaluate our approach on heterogeneous datasets extensively demonstrating its efficacy.

Performance degradation owing to data heterogeneity and low output interpretability are the most significant challenges faced by federated learning in practical applications. Personalized federated learning diverges from traditional approaches, as it no longer seeks to train a single model, but instead tailors a unique personalized model for each client. However, previous work focused only on personalization from the perspective of neural network parameters and lack of robustness and interpretability. In this work, we establish a novel framework for personalized federated learning, incorporating Bayesian methodology which enhances the algorithm’s ability to quantify uncertainty. Furthermore, we introduce normalizing flow to achieve personalization from the parameter posterior perspective and theoretically analyze the impact of normalizing flow on out-of-distribution (OOD) detection for Bayesian neural networks. Finally, we evaluated our approach on heterogeneous datasets, and the experimental results indicate that the new algorithm not only improves accuracy but also outperforms the baseline significantly in OOD detection due to the reliable output of the Bayesian approach.
\end{abstract}

\section{Introduction}

Distributed machine learning algorithms have attracted growing attention because of the success of data-driven artificial intelligence algorithms. However the realistic data all exist in silos, constrained by national privacy policies and enterprise business strategies, which cannot be shared unlike traditional distributed learning. As a paradigm of distributed learning, federated learning (FL) attempts to learn without direct access to user data, which has been favored by enterprises \cite{mcmahan2017communication}.

Real-world challenges have driven the industry not only to settle for collaborative training, but also to demand new requirements for federated learning. Constrained by data acquisition scenarios and complex environmental factors, real-world data is inherently heterogeneous, significantly impacting the performance of federated learning\cite{zhao2018fedwithnoniid}\cite{zhu2021fedonnoniidsurvy}\cite{li2022silosexperimental}. In response, personalized federated learning (PFL) has emerged, aiming to tailor a customized model for each client that performs well on their local data, while still benefiting from collaborative training\cite{kulkarni2020pfedsurvey}\cite{kairouz2021advances}. To establish a competent personalized federated learning framework, it is essential to strike a balance between the common knowledge shared across clients and the personalized components specific to each client.

Undoubtedly, the practical application of federated learning faces an additional obstacle related to the need for robustness and interpretability. This aspect is commonly known as trustworthy federated learning, particularly in industries like finance and healthcare. A feasible and commonly employed approach is to incorporate Bayesian approach, treating parameters as distributions to provide a measure of uncertainty in the output results\cite{mackay1995bnn}. By adopting this strategy, the model gains the capability to provide quantified output uncertainty, thus enhancing the algorithm's interpretability, addressing the concerns regarding the robustness and interpretability of the FL algorithms.

The introduction of BNN is not without its costs. Training BNN takes longer compared to point-estimate neural networks, as they need to learn the posterior distribution of parameters, and this difficulty increases superlinearly with the dimensions. Additionally, algorithms based on Markov Chain Monte Carlo (MCMC) require a large number of samples to approximate the posterior distribution\cite{welling2011bayesian}\cite{zhang2020cyclical}\cite{el2021federated}, making them less communication-efficient in a federated learning setting. Moreover, the phenomenon of cold posterior remains without a suitable theoretical explanation, which requires the introduction of additional hyperparameters during the training process\cite{wenzel2020good}. On the other hand, variational inference algorithms \cite{graves2011practical}\cite{blundell2015weight}\cite{dusenberry2020efficient} can reduce the number of parameters, but the posterior distribution obtained may not effectively approximate the true distribution. Indeed, some algorithms are based on understanding the relationship between the loss landscape and the posterior distribution. They use information from the loss landscape to efficiently approximate the posterior distribution\cite{maddox2019simple}\cite{izmailov2020subspace}\cite{daxberger2021laplace}. However, these algorithms, while efficient, are unable to eliminate the approximation loss entirely.

In this paper, we focus on the disparity between the obtained posterior distribution and the true posterior distribution. We introduce a novel personalized Bayesian federated learning framework equipped by posterior fine-tune (pFedPF) to aimed at eliminating the discrepancies between distributions. In addition, we integrate the current research on BNN and present a practical algorithm that incurs negligible additional storage, communication, and computational overheads. Our main contributions can be summarized as follows:
\begin{itemize}
    \item We introduce a novel and well-compatible framework that leverages posterior fine-tuning to eliminate the disparities between the approximate posterior and the true posterior.
    \item We have theoretically proven that our algorithm does not compromise the Out-Of-Distribution (OOD) detection capability of BNN. By examining the obtained upper bounds, we can observe that the algorithm's impact remains within an acceptable range even in the worst-case scenario.
    \item Through extensive experiments on a diverse set of heterogeneous data from the real world, we demonstrate the computational efficiency and robustness of our algorithm. Moreover, we have extensively compared numerous OOD algorithms across various datasets, and in most cases, our algorithm outperforms the baseline methods.
\end{itemize}

\section{Related Work}

% \textbf{Federated Learning.} FedAvg \cite{mcmahan2017communication}is a standard approach in Federated Learning which aggregates the client's model by averaging in each communication round. Compared to gradient-based approaches that require aggregation in each round, FedAvg can perform aggregation after the client has performed several local updates, and therefore it requires fewer communication rounds and is more efficient in communication. Many works have adequately investigated the effectiveness of FedAvg in terms of convergence of communication efficiency, but there are papers pointing out its limitations in the face of heterogeneous data\cite{zhao2018fedwithnoniid}. How to enhance the performance of Federated Learning under heterogeneous data is the hotspots in federated learning researches. FedProx adds proximal term to local update, which makes the trained model closer to the global model\cite{li2020fedprox}. SCAFFOLD uses control variables to fix the client drift in local update\cite{karimireddy2020scaffold}.

\textbf{Federated Learning.} FedAvg \cite{mcmahan2017communication}is a standard approach in Federated Learning which aggregates the client's model by averaging in each communication round. Compared to gradient-based approaches, FedAvg requires fewer communication rounds and is more efficient in communication. But there are papers that point out its limitations in the face of heterogeneous data\cite{zhao2018fedwithnoniid}. To address the performance decline arising from heterogeneity of data, Personalized Federated Learning algorithms believe that each client should have its own unique model. Some researches treat parts of neural networks as personalized layers. For instance, LG-FedAvg \cite{liang2020think} and FedPer \cite{arivazhagan2019federated} consider the feature extractor and classifier as personalized components, while FedBABU \cite{OhKY2022fedbabu} trains the feature extractor and classifier in two separate stages.

\textbf{Bayesian Neural Network.} Obtaining the posterior distribution of BNN primarily involves two main methods. One approach relies on the MCMC\cite{welling2011bayesian}\cite{chen2014stochastic}\cite{zhang2020cyclical}, which is computationally expensive but provides an approximation close to the true posterior. The other approach is based on variational inference\cite{graves2011practical}\cite{blundell2015weight}\cite{dusenberry2020efficient}, which has a lower computational overhead, but relies on approximation methods to estimate the posterior. Many research efforts have been dedicated to exploring more practical approaches to obtain the posterior distribution\cite{gal2016dropout}\cite{maddox2019simple}. Furthermore, research leverages the Laplace approximation to efficiently obtain the posterior distribution of network parameters\cite{daxberger2021laplace}.

% \textbf{Bayesian Federated Learning.} There are several works tying to introduce MCMC methods\cite{el2021federated}\cite{plassier2021dg}\cite{deng2021convergence}\cite{vono2022qlsd}\cite{kotelevskii2022fedpop}, but the majority still rely on approximating the posterior using parametric distribution. \citet{thorgeirsson2020probabilistic} approximate a Gaussian posterior distribution on the server, but their client model is deterministic. \citet{liu2021bayesian} uses KL divergence to find a global model that most closely resembles the client's model which is BNN, \citet{zhang2022personalized} adds a personalization component to the algorithm and give a theoretical proof. FedBe treats the client's model as a sampling of the global posterior\cite{chen2020fedbe}, and FedPa considers the global posterior to be equal to the product of local posteriors\cite{al2021fedpa}. \citet{yurochkin2019bayesian} uses Bayesian nonparametric learning to aggregate fully connected layers, and its subsequent work extended this result to convolutional layers\cite{wang2020federated}. \citet{achituve2021personalized} using Gausssian process as personalized classifier. \citet{kassab2022stein} using stein variational gradient descent to approximate posterior. Our methodology employing normalizing flow for posterior refinement to achieve personalization. This renders the parameterized posterior distribution more closely aligned with the actual posterior distribution, thereby enhancing the capacity for quantifying uncertainty. Our approach is adaptable to any parameterized approximation technique.

\textbf{Bayesian Federated Learning.} There are several works \cite{el2021federated}\cite{plassier2021dg}\cite{deng2021convergence}\cite{vono2022qlsd}\cite{kotelevskii2022fedpop}\cite{kassab2022stein} tying to introduce sample-based methods. But the majority still rely on approximating the posterior using parametric distribution. Parametric approximation algorithms\cite{thorgeirsson2020probabilistic}\cite{liu2021bayesian}\cite{zhang2022personalized}\cite{chen2020fedbe}\cite{al2021fedpa} often employ the Gaussian distribution to approximate the posterior, thus reducing computational and communication storage costs. However, the imprecision of the posterior approximation leads to suboptimal performance. There are works using Bayesian nonparametric learning to aggregate neural networks\cite{yurochkin2019bayesian}\cite{wang2020federated} and \cite{achituve2021personalized} using a Gausssian process as a personalized classifier. These endeavors involve modifications to either the client or server neural networks, rendering compatibility with other algorithms challenging and offering limited assistance to current researches.
Our algorithm incorporates the Post hoc approximation technique and an aggregation method based on KL divergence, thereby ensuring compatibility with contemporary research initiatives. Our methodology employs normalizing flow for posterior fine-tune to achieve personalization. This makes the parameterized posterior distribution more closely aligned with the actual posterior distribution, thus enhancing the capacity to quantify uncertainty.

\textbf{OOD Detection.}
Existing methodologies for detecting out-of-distribution (ODD) samples are predominantly based on two complementary approaches. The first perspective focused on post hoc. Particularly, \cite{hendrycks2016baseline} showed that a deep, pre-trained classifier typically exhibits a lower maximum softmax probability for anomalous examples compared to in-distribution examples. Consequently, such a classifier can effectively serve as a reliable OOD detector. Based on this work, \cite{liang2017enhancing} introduced the calibrated softmax score, ODIN, offering a refined approach to softmax probabilities. 
 \cite{lee2018simple} contributed by incorporating Mahalanobis distance into OoD detection, adding a statistical dimension to the process and \cite{liu2020energy} proposed an energy score method.  
 
 % \cite{xiao2020likelihood} introduced Likelihood Regret, and \cite{winkens2020contrastive} developed the Confusion Log Probability (CLP) score. 
 
 % Further advancements were made by \cite{lin2021mood} with an adjusted energy score, and more recently, \cite{sun2022out} with a k-th nearest neighbor (KNN) method and \cite{wang2022vim} with Virtual-logit Matching (ViM), broadening the scope of detection methodologies.

Another perspective emphasizes the importance of regularization during training, integrating OOD samples directly into the learning process. This approach is based on the use of real or synthetic OOD samples. Real OOD samples, typically derived from natural auxiliary datasets, have been explored by \cite{hendrycks2018deep}, \cite{mohseni2020self}, \cite{zhang2021fine}. However, the practical challenges of collecting real OOD samples, especially in resource-limited environments, have led to a growing interest in virtual synthetic OOD samples. This alternative avenue was explored by \cite{grcic2020dense} who trained a generative model to create synthetic OOD samples.
% and by \cite{jung2021standardized}, who proposed the detection of samples with different distributions by standardizing max logits. 
% Tack et al. (2020) and Sehwag et al. (2021) introduced contrastive learning methods that circumvent the need for real OOD samples. 
% Furthering this line of research, Du et al. (2022) proposed the VOS method for synthesizing virtual OOD samples based on the low-likelihood region of the class-conditional Gaussian distribution.

\section{Problem Setup}
Let us assume that there are $N$ clients, each with its own unique dataset, denoted $\mathcal{D}_{1}, \ldots, \mathcal{D}_{N}$. Here, $\mathcal{D}_{i}={(x_{n}, y_{n})}_{n=1}^{N{i}}$ represents the data labeled for the client $i$. Our model is indicated as $f_{w}\colon \mathcal{X}\to \mathcal{Y}$, where $w \in \mathbb{W}$ represents the model parameters, and we employ the loss function $\ell\colon \mathcal{X} \times \mathcal{Y} \to \mathbb{R}$. 

In the context of federated learning, we deal with multiple clients, and thus $w_{i}$ refers to the model weights of client $i$. Then the goal of PBFL can be formalized as follow
% \begin{equation}
%     \min\limits_{W}\{F(W) := \frac{1}{N} \sum_{k=1}^{N} \ell(w_i;x,y)\}
% \end{equation}
% where $W={w_1, w_2, \ldots, w_N}$ are the collection of all local models. Regarding PBFL, our primary focus lies in the distribution of parameters. Consequently, the optimization objective will differ in its specifics.
\begin{equation}
    \min\limits_{W}\{F(W) := \frac{1}{N} \sum_{k=1}^{N} \mathbb{E}_{w_i \sim p(w_i|\mathcal{D}_i)}[\ell(w_i;x,y)\}
\end{equation}

The posterior distribution is the most significant distinction between BNN and point-estimate neural networks. Accurate estimation of the posterior distribution will determine the predictive accuracy and uncertainty quantification performance of BNN\cite{wenzel2020good}\cite{izmailov2021bayesian}. Previous research has mainly focused on approximating the posterior distribution, our objective is to fine-tune the approximate posterior distribution to make it closer to the true posterior distribution.

\section{Preliminaries}
\subsection{Bayesian deep learning}
% The key distinction between BNN and point-estimate neural networks (NN) lies in how they treat the parameters of the neural network. BNN consider the neural network parameters as random variables $\theta \sim p(\theta)$, and thus, they do not aim to find a specific set of parameters. Instead, they focus on the posterior distribution $p(\theta|\mathcal{D})$ of the parameters. It is evident that this posterior distribution is highly complex and not easily parameterized. Therefore, when utilizing BNNs, we need to approximate the posterior distribution and employ Monte Carlo integration for certain marginalization computations.

 The uniqueness of Bayesian methods lies in introducing uncertainty into the parameters. Instead of considering a single model, we take into account all possible models. Therefore, the key to the success of Bayesian methods lies in how we obtain the likelihood of different models, which is essentially the process of obtaining the distribution of parameters.

\subsubsection{Posterior Of Bayesian Neural Network}

We can obtain approximate posterior estimates through the MCMC-based\cite{welling2011bayesian}\cite{chen2014stochastic}\cite{zhang2020cyclical} and VI-based methods\cite{graves2011practical}\cite{blundell2015weight}\cite{dusenberry2020efficient}. However, in federated learning, transmitting MCMC samples significantly increases communication overhead several times, while employing VI methods may lead to imprecise posterior estimates.

There are some approximation methods that can achieve training costs comparable to those of deterministic neural networks. For example, SWAG collects the training trajectories of a neural network to approximate a Gaussian distribution\cite{maddox2019simple}. Laplace approximation\cite{daxberger2021laplace} uses the parameters obtained from MAP (Maximum A Posteriori) training as the mean of a Gaussian distribution and calculates the covariance using the Hessian matrix at that point. 
\begin{equation}
\label{eq:laplace}
    w \sim \mathcal{N}(w_{\text{MAP}}, (\nabla_{w}^2 \mathcal{L}(\mathcal{D};w)|_{w_{\text{MAP}}})^{-1})
\end{equation}
These techniques provide computationally efficient ways to approximate the posterior distribution in Bayesian neural networks.

\subsubsection{Prediction}
Once we have obtained the posterior distribution, which represents the possibility of different parameter values, we can predict the label of new data by averaging over all these likely models
\begin{equation}
    p(y|x,\mathcal{D})=\int_{w} \underbrace{p(y|x,w)}_{\text{Data}} \underbrace{p(w|\mathcal{D})}_{\text{Model}}dw
\end{equation}
Due to the complexity of the likelihood and posterior, the integration involved in obtaining the analytical form of the solution is intractable. Therefore, we often utilize Monte Carlo integration to obtain an approximate solution.
\begin{equation}
    p(y|x, \mathcal{D}) \approx \frac{1}{M}\sum_{i=1}^{M} p(y|x, w^{(i)})
\end{equation}
where $w^{(i)}, i=1,...,M$ are samples which sampled from posterior $p(w|\mathcal{D})$

\subsection{Normalizing Flow}
Suppose we have a very complex distribution over $y$ and a simple distribution over $x$, we can utilize a transformation $T$ to build a connection between $y$ and $x$.
\begin{equation}
    y=T(x) \quad \text{where} \quad x \sim p_{x}(x).
\end{equation}
Base on the change of variable theorem, we can easily get the density of $y$:
\begin{equation}
    p_y(y) = p_x(x)|\det J_T(x)|^{-1} \quad \text{where} \quad x=T^{-1}(y). 
\end{equation}
Where $J_T$ is the Jacobian matrix. The transformation $T$ must be invertible and differentiable as well as it's invert transformation $T^{-1}$. These transformation are known as diffeomorphisms. 

Diffeomorphisms possess favorable properties, which allows us to combine them so that the resulting transformation remains a diffeomorphism. Given two transformation $T_1$ and $T_2$, their compositon defined as $T_2 \circ T_1$, we can calculate the inverse and Jacobian determinant as follow:
\begin{equation}
    (T_2 \circ T_1)^{-1} = T_1^{-1} \circ T_2^{-1}
\end{equation}
\begin{equation}
    \det J_{T_2 \circ T_1}(x) = \det J_{T2}(T_1(x))\cdot \det J_{T_1}(x)
\end{equation}
In practice, we can use multiple transformations chained together to achieve a sufficiently complex transformation, similar to a flow. Indeed, each transformation in a normalizing flow has its own parameters. Similar to other probabilistic models, we can perform parameter inference by minimizing the discrepancy between the flow-based model $p_y(y;\psi)$ and target distribution $p^*(y)$.

Common discrepancy measurement methods include KL divergence, f-divergence, integral probability metrics and so on. Since we will use it next, here we provide the definition of the reverse KL divergence:
\begin{equation}
\label{eq:reverse KL}
\begin{aligned}
    \mathcal{L}&=D_{KL}[p_y(y;\psi)||p^*_y(y)]\\
    &=\mathbb{E}_{p_x(x)}[\log p_x(x) - \log|\det J_{T}(x;\psi)|\\
    & \quad - \log p_y^*(T(x;\psi))]\\
\end{aligned}
\end{equation}
And we can estimate $\psi$ through:
\begin{equation}
    \begin{split}
    &\psi = \text{arg min}_{\psi} -\mathbb{E}_{p_x(x)}[\log|\det J_{T}(x;\psi)| \\
    & \quad + \log p_y^*(T(x;\psi))]\\
    \end{split}
\end{equation}

\subsubsection{Radial Flow}
The radial flow seeks to modify the distribution around a specific point by either radial contraction or expansion, and take the following form
\begin{equation}
    y=x + \frac{\beta(x-x_0)}{\alpha+||x-x_0||}
\end{equation}
where $\alpha$ and $\beta$ are the parameters and $\alpha > 0$

\section{Proposed Method}
In this work, we will propose a novel framework of personalized Bayesian federated learning from posterior distribution tuning perspective. Subsequently, we will provide a practical implementation of this framework, which, to some extent, mitigates the endogeneity limitations intrinsic to Bayesian methods.

As previously expounded, the communication overhead of MCMC algorithms relies on the number of samples, making the utilization of parameterized distribution for approximating the posterior distribution more advantageous in the realm of communication. Therefore we assume that the distribution of NN follows Gaussian which is commonly used\cite{blundell2015weight}\cite{chen2020fedbe}\cite{al2021fedpa}\cite{liu2021bayesian}.
\begin{equation}
    w_i \sim \mathcal{N}(\mu_i, \Sigma_i)
\end{equation}

Commonly, we may amalgamate a neural network feature extractor and classifier, an idea pervasive in current research\cite{tan2022fedproto}\cite{liu2022fedfr}\cite{xu2023personalized}. We take this notion into consideration. The feature embedding function $f_\theta:\mathcal{X} \to \R^d$ is the network parameterized by $\theta$ and $d$ represent the dimension of feature embedding. Given a data $x$, we can get the feature by $z=f_\theta(x)$, and using $g(z)$ which parameterized by $\phi$ to obtain the prediction. Our algorithm remains indifferent to which feature extractor and classifier among them assumes the role of the personalized layer. Our focus lies solely on applying the Bayesian approach to the classifier $\phi_i \sim \N(\mu_i, \Sigma_i)$, as its efficacy has been extensively demonstrated through extensive empirical or theoretical research\cite{kristiadi2020being}\cite{daxberger2021laplace}\cite{kristiadi2022posterior}, which allows us to further compress the additional communication overhead and computational burden. Then we can get the prediction as follow:
\begin{equation}
    p(y|x,\theta,\phi) = \int g(y|f_{\theta}(x),\phi) p(\phi|\D) d\phi
\end{equation}

To enhance the generality of our algorithm and facilitate its integration with current research, we can employ Laplace approximation and the SWAG algorithm for posterior distribution $p(\phi|\D)$ approximation\cite{maddox2019simple}\cite{daxberger2021laplace}. Both methods are compatible with SGD training. By doing so, our algorithm can be seamlessly combined with previous FL and PFL researches, while additionally providing the ability to quantify uncertainty.

In fact, the ability to quantify uncertainty using this approach is not completely exhaustive because the true posterior distribution $\tp(\phi|\D)$ is inherently complex, whereas we have utilized parameterized approximation methods. To address this limitation, we employ NF to fine-tune the approximated posterior distribution, thereby aiming to approach the true posterior. Let $T_{\psi_l}:\R^d \to \R^d$ be a diffeomorphism which parameterized by $\psi_l$ with inverse $T_{\psi_l}^{-1}$ and $d \times d$ Jacobian matrix $J_{T_{\psi_l}}$ for $l=1,\ldots,L$, Let $T_{\psi}$ denote $T_{\psi_L} \circ \cdots \circ T_{\psi_1}$ and $\psi := (\psi_l)_{l=1}^{L}$, then we can obtain the more precise posterior by
\begin{equation}
\label{eq:pf}
    \tp_\psi(\tilde{\phi}) = p(T_{\psi}^{-1}(\tilde{\phi}))\left|\prod_{l=1}^{L} \det J_{T_{\psi_l}^{-1}} (\tilde{\phi})\right|
\end{equation}

The parameter $\psi$ can obtained by Eq.\ref{eq:reverse KL}. However, it comes with a cost: the adjusted posterior distribution cannot be fully parameterized, and consequently, the uploaded distribution remains unchanged from its pre-tuned form. As for distribution aggregation, we will follow previous work by minimizing their KL-divergence\cite{liu2021bayesian}\cite{zhang2022personalized}.
\begin{equation}
\label{eq:bnn aggregation}
    \begin{aligned}
        \mu &= \sum_{i=1}^{N} \pi_i \mu_i\\
        \Sigma &= \sum_{i=1}^{N} \pi_i (\Sigma_i + \mu_i\mu_i^T - \mu\mu^T)\\
    \end{aligned}
\end{equation}
where $\pi_i=\frac{|\D_i|}{|\D|}$. If we consider the deterministic network as an equivalent to a Bayesian network with $0$ variance, then our approach will align with many contemporary researches, such as FedAvg.

\section{Theoretical Analysis}
\subsection{Out-Of-Distribution Detection Property}
In this section, we demonstrate that fine-tuning the posterior distribution of a  BNN using normalizing flows will not diminish its Out-of-Distribution (OOD) detection capabilities. At the beginning, we need to establish a clear definition of what constitutes possessing OOD capabilities; we use the result form \cite{hein2019relu} which points out why Relu network is overconfident.

\begin{theorem}[ReLU Overconfident]
Let $\mathbb{R}^d = \cup_{r=1}^{R}Q_r$ and $f|_{Q_r}(x)=U_r x + c_r$ be the affine representation in piecewise representation of the output of a ReLU network on $Q_r$. For almost any $x \in \mathbb{R}^n$ and $\forall \epsilon > 0$, there $\exists \delta > 0$ and class $i \in{1,...,k}$ such that $\text{softmax}(f(\delta x),i) \geq 1-\epsilon$, and, moreover, $\lim_{\delta \rightarrow \infty}\text{softmax}(f(\delta x),i)=1$, if $U_r$ does not contain identical rows for all $r=1,...,R$.
\end{theorem}

This theorem bears resemblance to the $\epsilon$-$\delta$-definition of limit. In essence, it conveys that as a new data deviate farther from the training data, the neural network's predictions tend to converge towards $1$. This is precisely why we refer to neural networks as being overconfident.

If we can establish that as samples deviate from the training data, the network's predictions decrease, or the upper bounds of its predictions decrease (eventually approaching a uniform distribution), then we can confidently assert that the network possesses OOD detection capabilities.

Here, we take the binary classification scenario as an example. First, let us define $\sigma(|f_w(x)|) = \max_{i\in{0,1}} p(y=i|x,w)$, our goal is proof $\lim_{\delta \rightarrow \infty} \sigma(f_w(\delta x)) \leq g(\theta) < 1$ where $\theta$ is controllable by design, and we can utilize it to constrain the model's confidence in distant regions.  

Fortunately, previous studies \cite{kristiadi2020being} have demonstrated that in the asymptotic regime far from the training data, and BNN can achieve a trust calibration approaching a uniform distribution by controlling the mean and variance of the weight parameters of ReLU neural network. 

% \begin{theorem}
% (All-layer approximation). Let $f_\theta : \mathbb{R} \rightarrow \mathbb{R}$ be a binary ReLU classification network parameterized by $\theta \in \mathbb{R}$ with $p\geq n$, and let $\mathcal{N}(\theta|\mu,\Sigma)$ be the Gaussian approximation over the parameters. Then for any input $x\in\mathbb{R}^n$,
% \begin{equation}
%     \lim_{\delta \rightarrow \infty} \sigma(|z(\delta x)|) \leq \sigma \left(\frac{||\mu||}{s_{min}(J)\sqrt{\pi/8\lambda_{min}(\Sigma)}}\right)
% \end{equation}
% where $u \in \mathbb{R}$ is a vector depending only on $\mu$ and the $n \times n$ matrix $J:=\frac{\partial u}{\partial \theta}|_{\mu}$ is the Jacobian of $u$ w.r.t. $\theta$ at $\mu$. Moreover, if $f_{\theta}$ has no bias parameters, then there exists $\alpha > 0$ such that for any $\delta > \alpha$, we have that
% $$
%     \sigma(|z(\delta x)|) \leq \lim_{\delta \rightarrow \infty} \sigma(|z(\delta x)|)
% $$
% \end{theorem}

\begin{theorem}[Approximation With NF]
\label{theorem:NF}
Let $f_w : \mathbb{R} \rightarrow \mathbb{R}$ be a binary ReLU classification network parameterized by $w \in \mathbb{R}^d$ with $d\geq n$, and let $\mathcal{N}(w|\mu,\Sigma)$ be the Gaussian approximation over the parameters. Then for any input $x\in\mathbb{R}^n$,
\begin{equation}
    \lim_{\delta \rightarrow \infty} \sigma(|z(\delta x)|) \leq \sigma \left(\frac{s_{\max}(J_{T})||\mu||}{s_{min}(J^{T})\sqrt{\pi/8\lambda_{min}(\Sigma)}}\right)
\end{equation}
where $u \in \mathbb{R}^n$ is a vector depending only on $\mu$, the $d \times n$ matrix $J:=\frac{\partial u}{\partial w}|_{\mu}$ is the Jacobian of $u$ w.r.t. $\theta$ at $\mu$, the $d \times d$ matrix $J_T$ is the Jacobian of normalizing flow $T$, $s(\cdot)$ means the singular value, and $\lambda(\cdot)$ means the eigenvalue. 
\end{theorem}

\begin{lemma}
Let $\{Q\}_{l=1}^{R}$ be the set of linear regions associated with the ReLU network $f:\R^n \to \R^k$. For any $x\in\R^n$ there exists an $\alpha \geq 0$ and $t \in\{1, \ldots, R\}$ such that $\delta x \in Q_t$ for all $\delta \geq \alpha$. Furthermore, the restriction of $f$ to $Q_t$ can be written as an addine function $U^Tx + c$ for some suitable $U\in \R^{n\times k}$ and $c\in R^k$
\end{lemma}

\textit{proof.} We employ a Taylor expansion of first order for the parameters of the neural network $f_w(x) \approx f_\mu(x) + J_f(x)\cdot(w - \mu)$ where $J_f(x)$ is the Jacobian matrix in the data $x$. We can use probit approximation to get a analytic approximation about output
\begin{equation}
    \label{eq:probit approximation}
    p(y=1|x,\mathcal{D})=\sigma(f_\mu(x)/\sqrt{1+\pi/8S(x)})
\end{equation}
where $S(x)=J_f(x)\Sigma J_f(x)^T$.

We follow the conclusion in \cite{hein2019relu}, stating that if $\delta$ is sufficiently large, then $\delta x$ belong to a linear region and we reformulate the linear region as an affine function form $u^Tx+c$. Since we ultilize the normalizing flow to fine-tune the posterior of $w$, assuming the output of the normalizing flow is $\tilde{w}$ and $T:\mathbb{R}^d \rightarrow \mathbb{R}^d$ is the diffeomorphism. We can obtain the Jacobian matrix as follow:
\begin{equation}
    \begin{split}
    J_f(\delta x) &= \left(\frac{\partial(\mu^T\delta x)}{\partial \tilde{w}} +  \frac{\partial c}{\partial \tilde{w}}\right) \frac{\partial \tilde{w}}{\partial w}\\
    &=\delta J^T J_{T^{-1}}(\tilde{w})x+ (\nabla_w c) J_{T^{-1}}(\tilde{w})
    \end{split}
\end{equation}
Put it back to Eq.\ref{eq:probit approximation}, and as $\delta \rightarrow \infty$, we have
\begin{equation}
    \begin{split}
    &\lim_{\delta \rightarrow \infty} p(y=1|\delta x, \mathcal{D})\\
    &\approx \frac{|u^T T^{-1}(\delta x) + c|}{\delta\sqrt{\pi/8(J^TJ_{T^{-1}}x)^T\Sigma(J^TJ_{T^{-1}}x)}}\\
    &\leq \frac{|u^T T^{-1}(\delta x) + c|}{\delta\sqrt{\pi/8\lambda_{min}(\Sigma)||J^TJ_{T^{-1}}x||^2}}\\
    &\leq \frac{|u^T T^{-1}(\delta x) + c|}{\delta s_{min}(J^T)s_{min}(J_{T^{-1}})\sqrt{\pi/8\lambda_{min}(\Sigma)||x||^2}}\\
    \end{split}
\end{equation}

\begin{lemma}
\label{lemma:ub_of_inverse}
    Let a diffeomorphism $T:\R^n \to \R^n$ denote a radial flow with input $a$ and output $x$ and $T^{-1}$ denote its reverse transformation, for every $z \in \R^n$ $T^{-1}(x)z^T \leq xz^T + \beta ||z^T||$ 
\end{lemma}

% \textit{proof.} Base on the definition of radial flow and $u=T^{-1}(x)$ we will have
% \begin{equation}
%     \begin{split}
%         x &= T^{-1}(x)+\frac{\beta(T^{-1}(x)-\gamma)}{\alpha + ||T^{-1}(x)-\gamma||} \\
%         &= T^{-1}(x) + \beta \hat{y}(T^{-1}(x))\\
%     \end{split}
% \end{equation}

% where $\gamma$ is the center point of radial flow and $\hat{y}(x)$ denotes the normalized $x-\gamma$. It is hard to get what $\gamma$ is, so we will use a random unit vector $\hat{y}$ to represent it. 
% \begin{equation}
%     \begin{split}
%         xz^T - \beta \hat{y}(T^{-1}(x))z^T &\eq T^{-1}(x)z^T\\
%         xz^T + \beta||\hat{y}||\ ||z^T|| &\geq T^{-1}(x)z^T\\
%         xz^T + \beta||z^T|| &\geq T^{-1}(x)z^T\\
%     \end{split}
% \end{equation}

Base on Lemma.\ref{lemma:ub_of_inverse}, we can have
\begin{equation}
    \begin{split}
        &\leq \frac{|u^T T^{-1}(\delta x) + c|}{\delta s_{min}(J^T)s_{min}(J_{T^{-1}})\sqrt{\pi/8\lambda_{min}(\Sigma)||x||^2}}\\
        &\leq \frac{|\delta u^Tx+\beta||u^T|| +c|}{\delta s_{min}(J^T)s_{min}(J_{T^{-1}})\sqrt{\pi/8\lambda_{min}(\Sigma)||x||^2}}\\
        &\leq \frac{||u||}{s_{min}(J^T) s_{min}(J_{T^{-1}})\sqrt{\pi/8\lambda_{min}(\Sigma)}}\\
    \end{split}
\end{equation}

As in $\delta \to \infty$, the part of the unit vector will go to $0$, regardless of the direction of the unit vector.

\begin{lemma}
    \label{lemma:S_inv}
    Let $A \in \R^{n \times n}$ and $A^{-1}$ denotes the inverse matrix, then $s_{min}(A) =s_{max}^{-1}(A^{-1})$
\end{lemma}
\textit{proof.} Using SVD, we have $A=USV^T$, the inverse matrix will have $A^{-1}=(V^{-1})^T (S^{-1})^T U^{-1}$, since $S$ is diagonal matrix $S=\text{diag}(s_1,s_2,\ldots,s_n)$, it inverse will be $S^{-1}=\text{diag}(\frac{1}{s_1},\frac{1}{s_2}, \ldots, \frac{1}{s_n})$, we can use the property that all singular value are positive, then $\min \{s_1, s_2, \ldots, s_n\} = 1/\max\{s_1^{-1}, s_2^{-1}, \ldots, s_n^{-1}\}$, which is $s_{min}(A) =s_{max}^{-1}(A^{-1})$

Base on Lemma.\ref{lemma:S_inv} we have 
\begin{equation}
    \begin{split}
        &\leq \frac{||u||}{s_{min}(J^T) s_{min}(J_{T^{-1}})\sqrt{\pi/8\lambda_{min}(\Sigma)}}\\
        &\leq \frac{s_{max}(J_{T})||u||}{s_{min}(J^T) \sqrt{\pi/8\lambda_{min}(\Sigma)}}\\
    \end{split}
\end{equation}

This demonstrates the impact of the introduced NF on the upper bound, which differs by an additional $s_{\max}(J_T)$ compared to the upper bound of the BNN. This illustrates the worst-case scenario, in which the NF actually steers the posterior closer to the true posterior, implying that a tighter bound could potentially be found. However, current results are sufficient to indicate that NF does not compromise the OOD detection capability of BNN.

% \begin{theorem}
% (Last-layer approximation). Let $g:\mathbb{R}^d \rightarrow \mathbb{R}$ be a binary linear classifier defined by $g(\phi(x)):=w^T\phi(x)$ where $\phi:\mathbb{R}\rightarrow \mathbb{R}^d$ is a fixed RELU network and let $\mathcal{N}(w|\mu,\Sigma)$ be the Gaussian approximation over the last-layer's weight. Then for any input $x\in \mathbb{R}^n$,
% \begin{equation}
%     \lim_{\delta \rightarrow\infty} \sigma(|z(\delta x)|) \leq \sigma\left(\frac{||\mu||}{\sqrt{\pi/8\lambda_{min}(\Sigma)}}\right)
% \end{equation}
% Moreover, if $\phi$ has no bias parameters, then there exists $\alpha>0$ such that for any $\delta \geq \alpha$, we have that
% $$
%     \sigma(|z(\delta x)|) \leq \lim_{\delta \rightarrow \infty} \sigma(|z(\delta x)|)
% $$
% \end{theorem}

% \subsection{Discussion}
% \subsection{Algorithm Complexity}
% \subsection{Posterior Distribution}
% \subsection{Communication Efficiency}

\begin{table*}[t]
\centering
\caption{The OOD detection result on different datasets. Dataset in he left side  is in-distribution dataset, the remaining dataset means OOD dataset. The results are in bold; the lower FPR95, higher AUROC and higher AUPR are better. It appears that our approach are better than the baseline in most of cases.}
\begin{tabular}{clccc|ccc|ccc}
\toprule
\multirow{2}{*}{ID dataset} & \multirow{2}{*}{method} &\multicolumn{3}{c}{SVHN}&\multicolumn{3}{c}{CIFAR100}&\multicolumn{3}{c}{TEXTURE}\\
\cmidrule(lr){3-5}\cmidrule(lr){6-8}\cmidrule(lr){9-11}
& & FPR95 & AUROC & AUPR & FPR95 & AUROC  & AUPR  & FPR95 & AUROC & AUPR\\

\midrule
\multirow{7}{*}{CIFAR10}
& Ours & \textbf{0.2856} & \textbf{0.9050} & \textbf{0.6832} & \textbf{0.4735} & \textbf{0.8491} & \textbf{0.5651} &
\textbf{0.5236} & \textbf{0.8383} & \textbf{0.5669} \\
& MSP & 0.4576 & 0.8485 & 0.5142 & 0.5901 & 0.8048 & 0.4292 &
0.7520 & 0.7259 & 0.3887 \\
& Energy & 0.4115 & 0.8406 & 0.4523 & 0.6849 & 0.7714 & 0.4507 &
0.8209 & 0.6807 & 0.3415 \\
& Entropy & 0.4119 & 0.8598 & 0.5027 & 0.5745 & 0.7806 & 0.4695 &
0.7345 & 0.7267 & 0.3827 \\
& Maxlogit & 0.4079 & 0.8537 & 0.4948 & 0.6931 & 0.7765 & 0.4616 &
0.8050 & 0.6926 & 0.3608 \\
& ODIN & 0.3784 & 0.8942 & 0.4523 & 0.6849 & 0.7714 & 0.4507 &
0.8209 & 0.6807 & 0.3415 \\
% Mahalanobis & 0.4687 & 0.8320 & 0.5355 & 0.7496 & 0.7301 & 0.3799 &
% 0.3907 & 0.9249 & 0.8101 \\
& MCD & 0.4314 & 0.8298 & 0.4558 & 0.5969 & 0.7883 & 0.4397 &
0.8180 & 0.7085 & 0.3645 \\
% VIM & 0.2293 & 0.9319 & 0.7617 & 0.5857 & 0.7877 & 0.4629 &
% 0.4800 & 0.9045 & 0.8004 \\
\midrule
& & \multicolumn{3}{c}{SVHN}&\multicolumn{3}{c}{CIFAR10}&\multicolumn{3}{c}{TEXTURE}\\
\midrule
\multirow{7}{*}{CIFAR100}
& Ours & \textbf{0.3211} & 0.8593 & 0.5081 & \textbf{0.4935} & \textbf{0.8249} & \textbf{0.4657} &
\textbf{0.5342} & \textbf{0.8153} & \textbf{0.4264} \\
& MSP & 0.4139 & 0.8469 & 0.5134 & 0.6345 & 0.7948 & 0.4490 &
0.7028 & 0.7200 & 0.3205 \\
& Energy & 0.3373 & \textbf{0.9008} & \textbf{0.6735} & 0.8725 & 0.6978 & 0.3493 &
0.8382 & 0.6361 & 0.2548 \\
& Entropy & 0.3884 & 0.8755 & 0.5909 & 0.6165 & 0.7946 & 0.4385 &
0.7025 & 0.7192 & 0.3142 \\
& Maxlogit & 0.3528 & 0.8983 & 0.6608 & 0.8566 & 0.7265 & 0.3834 &
0.8350 & 0.6560 & 0.2736 \\
& ODIN & 0.3772 & 0.8516 & 0.5232 & 0.9021 & 0.7109 & 0.4171 &
0.7552 & 0.7694 & 0.4096 \\
& MCD & 0.4302 & 0.8484 & 0.5077 & 0.5796 & 0.8044 & 0.4644 &
0.7815 & 0.7387 & 0.3329 \\

\midrule
& & \multicolumn{3}{c}{CIFAR10}&\multicolumn{3}{c}{CIFAR100}&\multicolumn{3}{c}{TEXTURE}\\
\midrule
\multirow{7}{*}{SVHN}
& Ours & \textbf{0.1633} & \textbf{0.9446} & \textbf{0.7768} & 0.2005 & \textbf{0.9457} & \textbf{0.7891} &
\textbf{0.5863} & \textbf{0.8879} & \textbf{0.7199} \\
& MSP & 0.2072 & 0.9271 & 0.71304 & \textbf{0.1928} & 0.9303 & 0.7082 &
0.7772 & 0.8506 & 0.6623 \\
& Energy & 0.1960 & 0.9409 & 0.7725 & 0.2397 & 0.9377 & 0.7730 &
0.7392 & 0.8355 & 0.6723 \\
& Entropy & 0.1962 & 0.9340 & 0.7462 & 0.2614 & 0.9331 & 0.7565 &
0.7804 & 0.8522 & 0.6784 \\
& Maxlogit & 0.1947 & 0.9358 & 0.7545 & 0.2273 & 0.9359 & 0.7650 &
0.7840 & 0.838 & 0.6728 \\
& ODIN & 0.3610 & 0.8924 & 0.6754 & 0.3398 & 0.9016 & 0.7085 &
0.9106 & 0.7857 & 0.5950 \\
& MCD & 0.1970 & 0.9294 & 0.7365 & 0.2496 & 0.9339 & 0.7629 &
0.7089 & 0.8617 & 0.6816 \\

\bottomrule
\end{tabular}
\label{tab:ood}
\end{table*}

\begin{table*}[t]
\centering
\caption{The final calibration result on different datasets. The lower NLL and ECE are better. Our approach exhibits superior performance in terms of ECE compared to the baseline. The algorithms suffixed with "pf" denote posterior fine-tune.}
\begin{tabular}{lccccccccc}
\toprule
\multirow{2}{*}{method} &\multicolumn{3}{c}{MNIST}&\multicolumn{3}{c}{FMNIST}&\multicolumn{3}{c}{CIFAR10}\\
\cmidrule(lr){2-4}\cmidrule(lr){5-7}\cmidrule(lr){8-10}
& ACC & NLL & ECE & ACC & NLL & ECE & ACC & NLL & ECE\\
\midrule
FedAvg & 98.66 & 0.0400 & 0.1996 & 87.74 & 0.3801 & 0.1818 & 77.35 & 0.6889 & 0.1463\\
FedPer & 97.91 & 0.1500 & 0.3215 & 93.13 & 0.2971 & 0.2534 & 85.75 & 0.7105 & 0.3382\\
LG-FedAvg & 96.27 & 0.23 & 0.3123 & 92.01 & 0.5283 & 0.3419 & 75.28 & 1.8049 & 0.4184\\
\midrule
FedAvg-pf & 98.88 & 0.1139 & \tb{0.1201} & 91.26 & 0.2766 & \tb{0.1094} & 83.93 & 0.4927 & \tb{0.1287}\\
FedPer-pf & 97.38 & 0.3509 & \tb{0.2308} & 93.51 & 0.3129 & \tb{0.1345} & 86.88 & 0.4573 & \tb{0.1860}\\
LG-FedAvg-pf & 94.66 & 0.4141 & \tb{0.2277} & 90.45 & 0.6821 & \tb{0.2781} & 74.48 & 1.2229 & \tb{0.2385}\\
\bottomrule
\end{tabular}
\label{tab:calibration}
\end{table*}

\section{Experiments}

\subsection{Setup}
% \subsubsection{Datasets} We focus on the image classification task and validate our algorithm on three common image datasets: MNIST, Fashion-MNIST (FMNIST), and CIFAR10. For OOD detection task, we will use the test sets of the three dataset mentioned above as the in-distribution dataset, and the OOD dataset are list below
% \begin{itemize}
%     \item MNIST: FMNIST, EMNIST, NOISE
%     \item FMNIST: MNIST, EMNIST, NOISE
%     \item CIFAR10: SVHN, NOISE
% \end{itemize}
% where NOISE dataset with $\delta=2000$ are adopted by \cite{kristiadi2020being}  and serves the purpose of assessing the model's ability to quantify uncertainty in the asymptotic regime far from the training data. We use LeNet-5 for the MNIST and FMNIST experiments and ResNet44 for CIFAR10 experiments. For NF, we use radial flow with $l=10$.
\subsubsection{Dataset} For the out-of-distribution (OOD) detection task, I will use CIFAR-10, CIFAR-100, and SVHN as in-distribution (ID) datasets, with each serving as an OOD dataset for the others. For experiments examining the impact of posterior fine-tuning on various federated learning algorithms, we will employ CIFAR-10, MNIST, and Fashion-MNIST as the training datasets. Additionally, we will incorporate the NOISE dataset with $\delta=2000$ into the OOD dataset. The NOISE dataset are adopted by \cite{kristiadi2020being}  and serves the purpose of assessing the model's ability to quantify uncertainty in the asymptotic regime far from the training data.

\subsubsection{Data Partitioning} Similar to the aforementioned paper\cite{karimireddy2020scaffold}\cite{xu2023personalized}, each client possesses two main classes, the rest being minor classes. The distinction between main and minor classes lies in the difference in the data volume. We ensure that each client has an equal amount of data, with a portion ($p$) uniformly distributed, and the remaining data evenly distributed among the main classes. To maintain the test set distribution identical to the training set, we employ the same method to split the test set for each client, treating it as their individual test data. In federated experiments, we set $p=0.2$, and every client have 1500 train data and 500 test data.

\begin{table*}[t]
\centering
\caption{The final OOD detection result on different datasets. In dataset column, the first dataset means in-distribution dataset, the second dataset means OOD dataset. The better results are in bold; the lower FPR95 and the higher AUROC are better. It appears that our approach are better than the baseline in most of cases.}
\begin{tabular}{l cc cc| cc cc |cc cc}
\toprule
\multirow{2}{*}{Dataset} &\multicolumn{2}{c}{FedAvg}&\multicolumn{2}{c}{FedAvg-pf}&\multicolumn{2}{c}{FedPer}&\multicolumn{2}{c}{FedPer-pf}&\multicolumn{2}{c}{LG-FedAvg}&\multicolumn{2}{c}{LG-FedAvg-pf}\\
\cmidrule(lr){2-3}\cmidrule(lr){4-5}\cmidrule(lr){6-7}\cmidrule(lr){8-9}\cmidrule(lr){10-11}\cmidrule(lr){12-13}
& AUROC & FPR95 & AUROC & FPR95 & AUROC & FPR95 & AUROC & FPR95 & AUROC & FPR95 & AUROC & FPR95\\
% & AUROC & FPR95 & AUROC & FPR95 & AUROC & FPR95 & AUROC & FPR95 & AUROC & FPR95 & AUROC & FPR95\\
\midrule
MNIST-FMNIST 
& .9796 & .5382 & .9814 & \tb{.3720} 
& \tb{.9649} & .5878 & .9385 & \tb{.3775} 
& .9080 & .7031 & .8954 & \tb{.4601} \\
MNIST-EMNIST 
& .9058 & .6873 & \tb{.9538} & \tb{.4485} 
& .8950 & .7180 & \tb{.9052} & \tb{.4223} 
& .8632 & .7694 & .8613 & \tb{.4782} \\
MNIST-NOISE 
& .0596 & .9798 & \tb{.8223} & \tb{.6136} 
& .2809 & .9917 & \tb{.9495} & \tb{.3810} 
& .2435 & .9719 & \tb{.8935} & \tb{.4860} \\
\midrule
FMNIST-MNIST 
& .6464 & .7294 & \tb{.8865} & \tb{.5351} 
& .7364 & .8004 & \tb{.9128} & \tb{.4629}
& .7754 & .8146 & \tb{.8526} & \tb{.4682} \\
FMNIST-EMNIST 
& .5658 & .7530 & \tb{.8370} & \tb{.6005}
& .6760 & .8198 & \tb{.8757} & \tb{.5343} 
& .6913 & .8408 & \tb{.7847} & \tb{.5188}\\
FMNIST-NOISE 
& .0299 & .7994 & \tb{.8694} & \tb{.5513}
& .1395 & .8921 & \tb{.8867} & \tb{.4939}
& .2282 & .9150 & \tb{.7400} & \tb{.4983}\\
\midrule
CIFAR10-SVHN 
& .5081 & .6831 & \tb{.8306} & \tb{.5598} 
& .7746 & .7102 & \tb{.8247} & \tb{.6070}
& .6574 & .7907 & \tb{.6713} & \tb{.5844} \\
CIFAR10-NOISE 
& .0 & .7088 & \tb{.7716} & .6708 
& .1240 & .8389 & \tb{.9845} & \tb{.2882} 
& .0371 & .8281 & \tb{.8754} & \tb{.5034}\\
\bottomrule
\end{tabular}
\label{tab:ood2}
\end{table*}

\subsubsection{Metrics} In the realm of conventional federated learning tasks, the focus is typically on test accuracy. In order to assess the algorithm's robustness, we take two additional aspects into consideration: calibration and out-of-distribution (OOD) detection. For calibration performance, we will utilize Negative Log-Likelihood (NLL) and Expected Calibration Error (ECE). To evaluate the OOD-detection performance, we will use the Area Under the Receiver Operating Characteristic curve (AUROC), false-positive rate at $95\%$ true-positive rate (FPR95) and the area under the PR curve (AUPR). In some constrained spaces, we may disregard metrics where performance differences are minor or lack representativeness.

% \subsubsection{Metrics} In the realm of conventional federated learning tasks, the focus typically rests upon the test accuracy. As the inaugural exploration into quantifying the robustness and uncertainty quantification capabilities of personalized federated learning, we are poised to introduce metrics that gauge the calibration and out-of-distribution (OOD) detection proficiency of the model. Of course, our research centers around personalized models, hence the ensuing values discussed are derived by averaging the values across all client models. 

% For calibration performance, we will utilize Negative Log-Likelihood (NLL) and Expected Calibration Error (ECE). To evaluate the OOD-detection performance, we will use the Area Under the Receiver Operating Characteristic curve (AUROC) and false-positive rate at $95\%$ true-positive rate(FPR95). For over-confidence we will use mean maximum predictive confidence.

\subsubsection{Baselines} In addition to the FedAvg algorithm \cite{mcmahan2017communication}, we will compare the personalized federated learning approaches of FedPer \cite{arivazhagan2019federated}, which set the classifier as a personalized layer, and LG-FedAvg \cite{liang2020think}, which set the feature extractor as a personalized layer. By employing these two algorithms as baselines and contrasting their outcomes with the posterior fine-tuned versions, we can discern whether there exists a disparity in the efficacy of our approach in the context of personalized and common layers. This would elucidate the universality of the algorithm. Our method are integrate posterior fine-tune version.
For OOD tasks, our algorithm is compared to other post hoc methods such as MSP\cite{hendrycks2016baseline}, Energy\cite{liu2020energy}, Entropy and Maxlogit. As well as the methods which need extra processing like ODIN \cite{liang2017enhancing} and MCD\cite{gal2016dropout}.

\subsubsection{Implementation details} For the FL setting, we assume that there are 20 clients and the local epoch is set to $E=5$. All clients participate in each communication round, and the total number of communication rounds is set to $T=120$ for the CIFAR10 experiments and $T=80$ for the MNIST experiments and FMNIST experiments. At this point, the accuracy improvement of all algorithms is negligible. During local training, we use SGD as optimizer with a learning rate of $\eta = 0.01$, momentum of 0.9, and weight decay of 5e-4. The batch size is fixed at $B=128$. For the PF version of the algorithm, we employ Laplace approximation to acquire the posterior distribution, and utilizing radial flow with length $l=10$ for posterior fine-tuning. Comparisons of OOD algorithms are all based on the FedAvg-CIFAR10 setting.

We use LeNet-5 for the MNIST and FMNIST experiments and ResNet44 for CIFAR10 experiments. For the OOD task, to facilitate better comparisons, we will use the widely employed network in OOD research, namely WideResNet40, as our network architecture.

\begin{figure}[htbp]
\centering
\includegraphics[width=0.99\columnwidth]{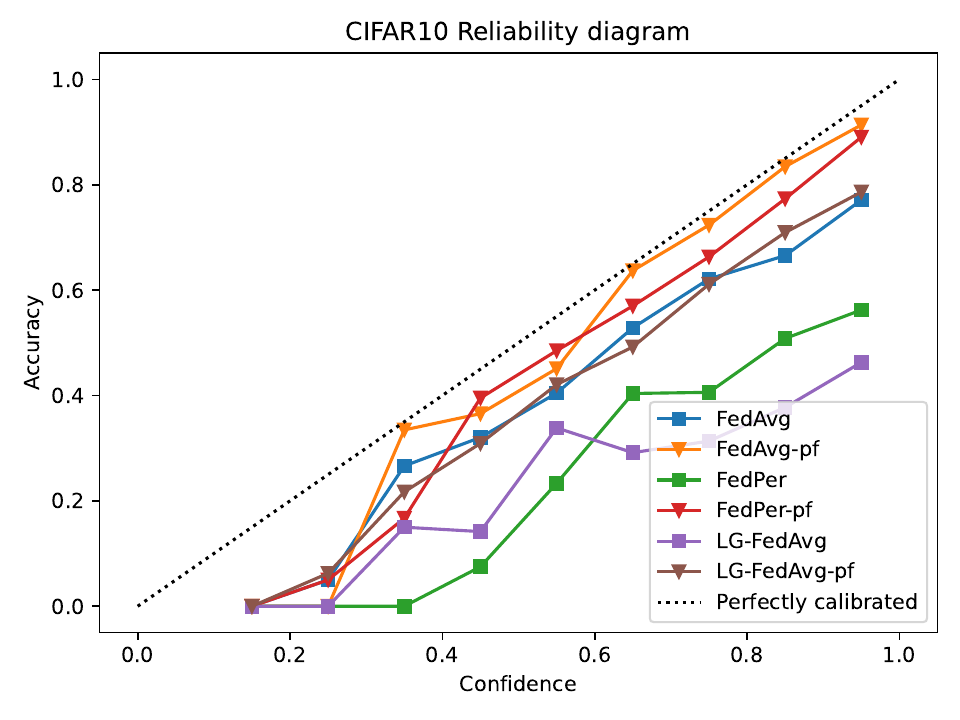} 
\caption{The reliability diagram of all algorithm in CIFAR10 dataset. This signifies that our approach will rectify the issue of model overconfidence.}
\label{fig:rd}
\end{figure}

\subsection{OOD Comparison In Federated Learning}

Our algorithm's OOD capability comes from its reliable outputs. We have processed our algorithm's outputs similarly to the MSP method. The results of the OOD comparisons are listed in Table.\ref{tab:ood}, where it can be seen that our algorithm outperforms the baseline in most cases. The significant improvement over MSP is noteworthy, indicating the advantages of more reliable output.

\subsection{The Influence of Posterior Fine-tune In Different algorithm}
\subsubsection{Calibration}
ECE is the difference in expectation between confidence and accuracy. If ECE equals $0$, that means that we can trust the result of the model. The results are listed in Table.\ref{tab:calibration}. The ECE of our algorithm exhibits satisfactory performance across different federated learning algorithms, indicating the enhanced reliability of our algorithm's outputs.

Reliability diagrams provide a visual understanding of model calibration. We present the Reliability Diagram for the CIFAR10 dataset in Fig.\ref{fig:rd}, with similar trends observed in the diagrams for other datasets. It is apparent that our approach is more closely aligned with a state of being perfectly calibrated.

% \subsubsection{Performance Comparison}
% These enhancements are not without their trade-offs. In terms of predictive accuracy, our algorithm may lag behind the baseline. In most cases, it aligns closely with the baseline, with differences within a range $2\%$ within an acceptable scope. It can be asserted that despite a slight deficit in predictive capability, our algorithm and the baseline are at a comparable level in terms of predictive accuracy.

% \begin{table}[t]
% \centering
% \caption{The final test accuracy on different datasets under non-IID settings. The The accuracy of the algorithm after posterior fine-tune is on par with that of the original algorithm.}
% \label{tab:accuracy}
% \begin{tabular}{llll}
% \toprule
% method &
% MNIST&
% FMNIST &
% CIFAR10\\
% \midrule
% FedAvg & 98.66 & 87.74 & 77.35\\
% FedPer & 97.91 & 93.13 & 85.75 \\
% LG-FedAvg & 96.27& 92.01 & 75.28\\
% \midrule
% FedAvg-pf & 98.88 & 91.26 & 83.93\\
% FedPer-pf & 97.38 & 93.51 & 86.88 \\
% LG-FedAvg-pf & 94.66 & 90.45 & 74.48\\
% \bottomrule
% \end{tabular}
% \end{table}

\subsubsection{OOD detection}
In the context of the OOD detection task, we have introduced the NOISE dataset to substantiate our earlier assertion that the use of NF still averts issues of overconfidence in the asymptotic regime.

As presented in Table.\ref{tab:ood}, Our method demonstrates strong performance in most metrics, particularly excelling in the FPR95 metric where it consistently exceeds the baseline. As for the AUROC metric, the results vary, with successes and setbacks on both sides, when considering the average, our approach significantly exhibits greater excellence. 
% Regarding the issue of model overconfidence, as depicted in Table.\ref{tab:confidence}, our method consistently surpasses the baseline comprehensively.

\subsection{Ablation Studies}
We have employed Laplace approximation and normalizing flow for posterior inference and fine-tune. We illustrate this with the example of the FedAvg algorithm on the CIFAR-10 dataset (similar trends are observed in other experiments as well). The results are presented in Table.\ref{tab:ablation}. The use of Laplace approximation can slightly enhance the performance of the original algorithm. Additionally, employing posterior fine-tuning techniques further strengthens performance. Moreover, as the number of NF layers increases, the performance improvement becomes more pronounced.

\begin{table}[t]
\centering
\caption{Ablation study. LA denotes Laplace approximation; NF denotes Normalizing Flow, the number after NF means the length of NF.}
\label{tab:ablation}
\begin{tabular}{llllll}
\toprule
method &
ACC$\uparrow$&
ECE$\downarrow$ &
NLL$\downarrow$ &
FPR95$\downarrow$ &
AUROC$\uparrow$ \\
\midrule
FedAvg & 77.35 & 0.1462 & 0.6830 & 0.6218 & 0.7743\\
\midrule
LA & 77.29 & 0.1298 & 0.6794 & 0.5851 & 0.7977\\
LA-NF-1 & 78.04 & 0.1327 & 0.6608 & 0.5862 & 0.8005\\
LA-NF-3 & 79.85 & 0.1329 & 0.6649 & 0.5807 & 0.8072\\
LA-NF-5 & 81.25 & 0.1289 & 0.5668 & 0.5729 & 0.8180\\
LA-NF-10 & 83.93 & 0.1287 & 0.4927 & 0.5598 & 0.8306\\
LA-NF-20 & \tb{87.62} & \tb{0.1247} & \tb{0.3973} & \tb{0.5402} & \tb{0.8483}\\
\bottomrule
\end{tabular}
\end{table}

\section{Conclusion}
In this paper, our focus lies in an unexplored realm, previously unattended to by predecessors, the distinction between the global posterior and the authentic client posterior. We employ NF to accomplish personalized posterior distributions, and by amalgamate post hoc approximation techniques, our method endows current non-Bayesian federated learning algorithms with the capacity to quantify uncertainty. We substantiate the efficacy of the algorithm both theoretically and empirically.

\bibliographystyle{IEEEtran}
\bibliography{ref}

\end{document}